# Assessing the Impact of the Quality of Textual Data on Feature Representation and Machine Learning Models

Tabinda Sarwar[1*+], Antonio Jose Jimeno Yepes[1+], and Lawrence Cavedon[1]

[1]*School of Computing Technologies, RMIT University, Melbourne, Australia*

**Corresponding author: tabinda.sarwar@rmit.edu.au*

[+]*Tabinda Sarwar and Antonio Jose Jimeno Yepes contributed equally to this research.*

## Abstract

**Background:** Data collected in controlled settings typically results in high-quality datasets. However, in real-world applications, the quality of data collection is often compromised. It is well-established that the quality of a dataset significantly impacts the performance of machine learning models. In this context, detailed information about individuals is often recorded in progress notes. Given the critical nature of health applications, it is essential to evaluate the impact of textual data quality, as any incorrect prediction can have serious, potentially life-threatening consequences.

**Objective:** This study aims to quantify the quality of textual datasets and systematically evaluate the impact of varying levels of errors on feature representation and machine learning models. The primary goal is to determine whether feature representations and machine learning models are tolerant to errors and to assess whether investing additional time and computational resources to improve data quality is justified.

**Methods:** A rudimentary error rate metric was developed to evaluate textual dataset quality at the token level. Mixtral Large Language Model (LLM) was used to quantify and correct errors in low-quality datasets. The study analyzed two healthcare datasets: the high-quality MIMIC-III public hospital dataset (for mortality prediction) and a lower-quality private dataset from Australian aged care homes (for depression and fall risk prediction). Errors were systematically introduced into MIMIC at varying rates, while the ACH dataset quality was improved using the LLM. Feature representations and machine learning models were assessed using the area under the receiver operating curve.

**Results:** For the sampled 35,774 and 6,336 patients from the MIMIC and ACH datasets respectively, we used Mixtral to introduce errors in MIMIC and correct errors in ACH. Mixtral correctly detected errors in 63% of progress notes, with 17% containing a single token misclassified due to medical terminology. LLMs demonstrated potential for improving progress note quality by addressing various errors. Under varying error rates (5% to 20%, in 5% increments), feature representation performance was tolerant to lower error rates (<10%) but declined significantly at higher rates. This aligned with the ACH dataset's 8% error rate, where no major performance drop was observed. Across both datasets, TF-IDF outperformed embedding features, and machine learning models varied in effectiveness, highlighting that optimal feature representation and model choice depend on the specific task.

**Conclusions:** The study revealed that models performed relatively well on datasets with lower error rates (<10%), but their performance declined significantly as error rates increased (≥10%). Therefore, it is crucial to evaluate the quality of a dataset before utilizing it for machine learning tasks. For datasets with higher error rates, implementing corrective measures is essential to ensure the reliability and effectiveness of machine learning models.

## 1. Introduction

Electronic health records (EHRs) represent rich information capturing the heterogeneous aspects of an individual's health such as medical history, vital signs, prescriptions, laboratory tests, imaging reports, and treatment plans. Because of this, EHRs have been now widely adopted in data-driven treatment approaches where data mining and machine learning approaches have been applied for clinical applications such as disease prediction, precision medicines, diagnosis, phenotype extraction and disease tracking [1–4]. Generally, a subset of the EHRs is recorded in a structured format e.g., vital signs and laboratory tests as these tend to look for specific characteristics of a patient. However, a major part of the EHRs is in the form of free text notes. This includes, but is not limited to, doctors' notes, disease symptoms, nurses' observations, findings from radiologists, and any adverse reactions to medication. This textual data, which encapsulates critical and complex information, has been extensively utilized in data mining and machine learning.

Textual data, irrespective of the field of medicine and health, holds immense importance. In our daily lives, much of the information we receive comes from textual sources, such as news articles, discussion forums, web pages, and research papers. Additionally, transcription has become widely adopted to convert audio into text. Numerous natural language processing (NLP) techniques have been explored in the literature for data mining and predictive tasks [5–8]. The introduction of the transformer model [9] marked the beginning of a new era in deep learning, enabling the simultaneous processing of text sequences and capturing word context more effectively [10–13]. This led to significant improvements in efficiency and performance compared to traditional NLP models. Subsequently, this progress paved the way for the development of many models now referred to as Large Language Models (LLMs) [14–17]. A large number of LLMs exist in literature, each striving to outperform its predecessor, though this often comes at the cost of increased computational power.

LLMs were initially developed using general domain datasets (e.g., Wikipedia, BooksCorpus, public GitHub repositories, etc.), which are not representative of biomedical text [17, 18].The biomedical corpus differs significantly from the general corpus due to its specialized vocabulary and concepts, limiting the application of general LLMs in the biomedical and clinical domains. Recognizing this limitation led to the development of domain-specific LLMs, such as BioBERT [19], ClinicalBERT [20], PubMedBERT [21], BioMed-RoBERTa [22], SciBERT [23], MedCPT [24] and GatorTron [25]. These models were trained on biomedical domain corpora (PubMed abstracts, and articles from PMC and Semantic Scholar) and open-source hospital datasets (MIMIC-III [26], and i2b2 [27, 28]), except GatorTron which utilized medical information from the private University of Florida Health Integrated Data Repository.

The data sources used for training biomedical or clinical versions of LLMs are considered high-quality datasets, with minimal grammatical, typological, and spelling errors. However, data collected from real-life applications often contain many such mistakes, including informal language use and non-standardized abbreviations. It is well-established that data quality significantly impacts the performance of machine learning and deep learning models [29–33]. Identifying errors in structured datasets is relatively straightforward due to their standardized and objective representation. In contrast, assessing the quality of textual data is more complex due to its subjective and unstructured nature. As textual data is increasingly used in daily routines, it is crucial to evaluate the impact of its quality on machine learning applications, especially in biomedical and clinical contexts. It should be noted that the quality of structured data has been extensively studied in the literature, along with its impact on machine learning algorithms [29, 30]. However, the quality of unstructured textual data has not been thoroughly evaluated, particularly considering its challenging and subjective nature [29, 33].

In this study, we aim to systematically assess the impact of textual dataset quality on the performance of predictive machine learning models. We begin by defining and categorizing different types of errors that can occur in textual datasets. Artificial errors were then systematically introduced and controlled within the datasets to compare model performance across varying error rates. The study leverages both traditional (TD-IDF) and neural network (word2vec and BERT) embedding-based models to determine which models are more affected by poor-quality data. The generated embeddings were then used for the predictive tasks using state-of-the-art machine learning models (logistic regression, random forests and XGBoost). We utilized publicly available (MIMIC-III [26]) and private (outsourced from aged care homes) healthcare datasets. The study focuses on biomedical data, emphasizing the importance of accuracy in this domain, as any error or bias in the data or models can have significant implications for health and patient outcomes. This is the first study to systematically evaluate the performance of NLP feature representation and machine learning models under varying levels of textual data quality.

## 2. Related Work

Significant literature exists on assessing the quality of structured data [31, 34–37], including aspects of EHRs [3, 38–40]. These studies evaluated data quality across various dimensions, such as completeness, accuracy, integrity, validity, and consistency. These dimensions further addressed data issues such as outliers, class imbalance, missing values, correctness, realistic values, and duplicates. Due to its objective nature, structured data is easier to assess; for example, a patient's temperature is represented by a single objective value, either in Fahrenheit or Celsius, with a specific realistic range. However, assessing the quality of textual data is more challenging due to its subjective nature.

Several preliminary studies have explored methods for assessing the quality of textual data. Kiefer [41] suggested the following indicators for text data quality: 1) percentage of abbreviations, 2) percentage of spelling mistakes, 3) lexical diversity, 4) percentage of uppercased words, and 5) percentage of ungrammatical sentences. Larson et al. [42] proposed a pipeline for dialogue systems to effectively detect two types of textual outliers: 1) errors: sentences that have been mislabeled that can have a detrimental impact on the classification model, and 2) unique samples: sentences that differ structurally from most others in the data but can enhance the model's robustness. A few studies have not directly assessed the quality of textual data but have instead focused on understanding complex text classification tasks [43, 44], improving models to generate diverse textual outputs and reduce the production of generic responses [45], and detecting noise in labels [46–49].

In the context of machine learning tasks, there is limited research on textual dataset quality [50]. Swayamdipta et al. [51] proposed "Data Maps" to identify textual samples that are easy to learn, hard to learn, and ambiguous for classification models, suggesting that hard-to-learn instances often correspond to data errors. Ribeiro et al. [52] developed a checklist for testing NLP models with respect to various aspects, including vocabulary, taxonomy, robustness to typos and irrelevant changes, named entity relationships, fairness, temporality, negation, coreference, semantic role labelling, and logic. Colavito et al. [53] primarily assessed the quality of labels in classification tasks using pre-trained language models (BERT, ALBERT, and RoBERTa).

A subset of the mentioned studies attempted to identify textual quality errors, but none conducted a systematic study to evaluate the value of investing extra effort in improving data quality and its association with the performance of machine learning models. Machine learning tasks involving medical textual datasets are more critical than those using other types of textual data (e.g., social media, news articles) because they have a direct impact on human health and life. Therefore, it is essential to systematically evaluate the effects of erroneous low- and high-quality medical textual datasets on machine learning tasks in the healthcare domain.

## 3. Methods

In this study, we conducted a systematic analysis of the impact of textual dataset quality on NLP feature representation and machine learning models. The study's framework is illustrated in Figure 1. Briefly, we utilized medical datasets, both publicly available and private, representing high-quality (with no errors) and low-quality (with errors) datasets (Section 3.1). We developed a pipeline to introduce errors into the clean data (Section 3.2). LLM (specifically, Mixtral) was then employed to process the low-quality data, correcting and removing errors. Both the processed and raw text datasets were subsequently used to generate feature vectors for classification tasks (Section 3.4). The outcomes for low- and high-quality datasets were systematically evaluated (Section 3.4) to assess the impact of errors on the performance of machine learning models.

### 3.1. Datasets

**MIMIC Dataset**

We used MIMIC-III [26] publicly available medical database, which consists of de-identified health-related data of over 40,000 ICU patients in the Beth Israel Deaconess Medical Center in Boston, Massachusetts, between 2001 and 2012. The database contains comprehensive information about the patient, including demographics, medications, diagnoses, medical procedures, caregiver notes, and admission and discharge summaries. As the study aimed to assess the impact of the textual dataset, we limited the data selection to PATIENTS, ADMISSIONS, and NOTEEVENTS tables of the MIMIC-III database.

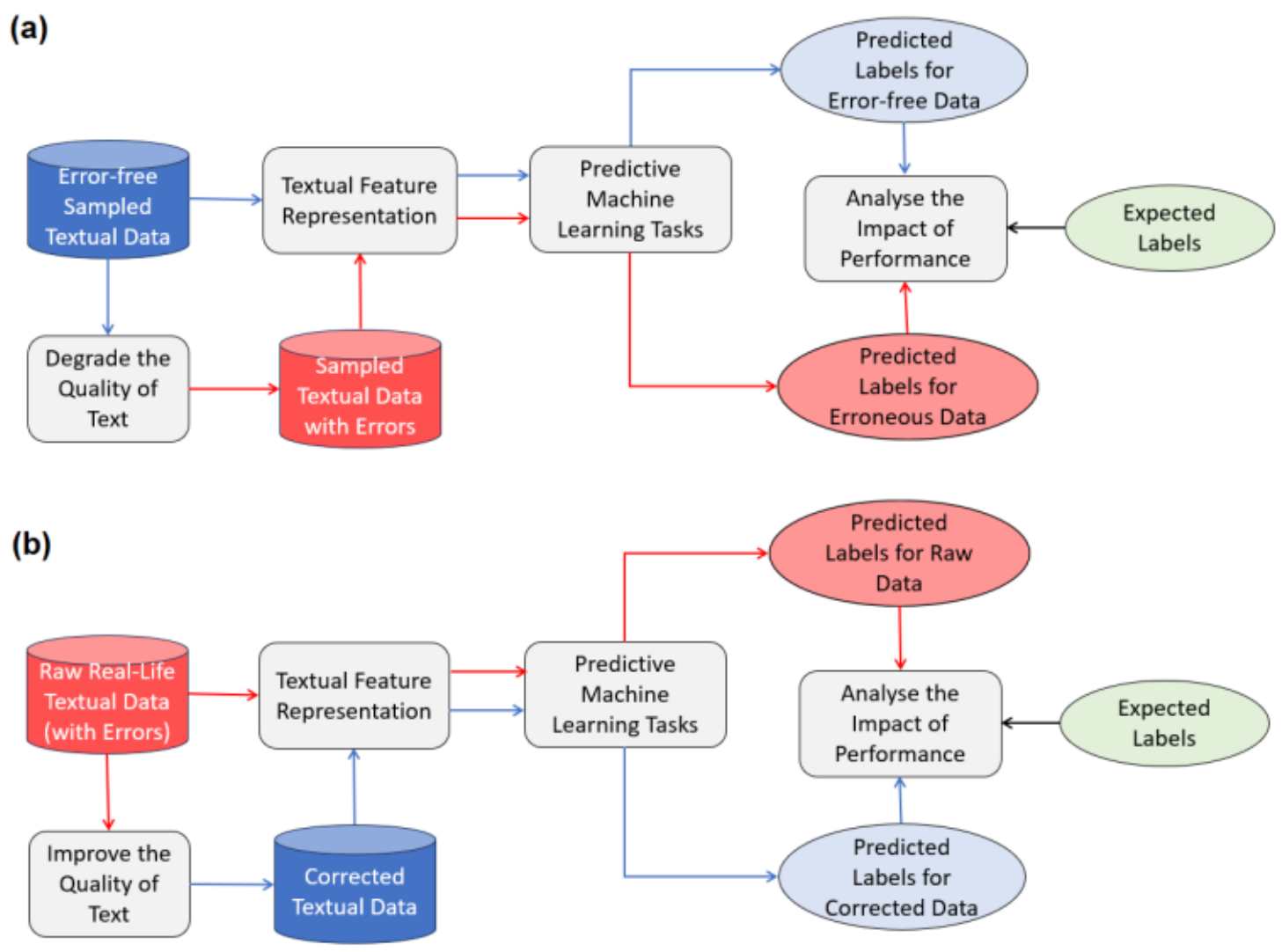

**Figure 1:** The framework for evaluating the impact of the quality of textual datasets on machine learning models. This involves (a) systematically degrading the high-quality data and (b) improving the low-quality data for the NLP feature representation and predictive models.

## Dataset from Aged Care Homes

Australian Aged Care Homes (ACHs) regularly record resident data to comply with the quality standards set by the Australian Aged Care Quality and Safety Commission (AACQSC). This data captured multiple aspects of residents' lives, including but not limited to demographics, medical history, medications, mobility, bowel movements, dietary requirements, and fall risk. A significant portion of the data consisted of progress notes—free-text entries that document the daily progress or condition of residents. Nursing staff (both caregivers and registered nurses) routinely record the health status of elderly residents, often incorporating input from general practitioners.

The anonymized operational electronic health records (EHRs) from ACHs, operated by the Royal Freemasons Benevolent Institution, based in New South Wales, Australia was acquired for this study. The dataset corresponded to 26 ACHs that had a total capacity of 1160 beds from 1 January 2010 to 1 January 2021. This dataset was acquired through Telstra Health, whose Clinical Manager tool is installed in over 320 ACHs for recording EHRs. Although the dataset comprised two broad categories, unstructured data (progress notes) and structured data (observation charts and assessment forms), we utilized only the progress notes, incident reports, medical history and demographic details for this study.

## 3.2. Data Processing

To assess the impact of textual data quality on feature representation and machine learning models, datasets containing both erroneous and error-free text notes were needed to represent low- and high-quality data, respectively. To achieve this, we adopted two strategies described in Sections 3.2.3 and 3.2.4. Firstly, this involved defining and quantifying the textual errors in the dataset (Sections 3.2.1 and 3.2.2).

### 3.2.1. Types of Errors

The progress noted from MIMIC was deemed as an error-free dataset, as it has been extensively used in the developing models and techniques in the medical and healthcare domain and no textual mistakes have been reported by any study yet. On the other hand, we observed many errors in the progress notes in ACHs dataset, which included: missing white-space between words, grammatical errors, typological mistakes, spelling mistakes, and informal abbreviations (e.g., "btw" instead of "between", and "hwr" instead of "however"). Examples of these mistakes are reported in Table 1.

**Table 1:** Sample of errors found in the textual dataset acquired from aged care homes

| Types of Error | Textual samples |
|---|---|
| Spelling Mistakes | "Staff spoke to XYZ regarding her eye lashes. XYZ to arrange a time with **optomristrist**"<br>"Resident has had very little sleep **overnite**"<br>"Resident was i/c of urine when staff went to check on her after her bed **sencer** went off" |
| Typological Mistakes | "Resident found sitting in chair on 0400hrs round check. Was **tioleted** and directed back to bed. Given water on both **occassions**"<br>"Assisted x 2 staff with ADLs and personal hygiene. **Transfered** with stand-up **lifer**"<br>"Resident later **appologised** to staff for her behaviour and said that it **wouldnt** happen again." |
| White-space Mistakes | "Resident sitting in the lounge area with fellow **residents,while** sitting Resident was picking at her scores on her **legs,when** staff asked Resident not to pick at them ."<br>"Resident buzzed around **0900hrs,staff** entered room to find Resident laying on her bed feeling **unwell.Resident** stated that she felt dizzy and wanted to go to the **toilet.Staff** assisted Resident to the **toilet,Resident** had her BO and remains in her room, staff to monitor throughout day."<br>"Antibiotics **conti nue** for infected toe - dressing remains intact this shift." |
| Grammatical Mistakes | "Resident was found incontinent of faeces while **sitted** in lounge chair."<br>"She **putted** toilet paper on the floor, she **is look** so confused."<br>"XYZ **has participate** with mainstream activities during the day."<br>"The resident **didnot have eat** much for dinner, she had a **cup tea**, **few spoonful** of soup and main meal." |
| Informal Abbreviations | "He was walking with his knee almost bended **coz** he was very tired from wandering."<br>"XYZ given. resident being reviewed by GP **tmw** for her pain and a suitable pain regime."<br>"Resident came down for **bfast** and ate all of her maeal." |
| Miscellaneous | "Left leg **swollen,in,two areas,c.m. shown,daughter rang,will** come in and take her mum up to Hospital, to have it checked."<br>"Post lunch staff **intervention/ s** required as resident became verbally distruptive w/ another resident."<br>"XYZ is also walked to crafts,and atttend as many group activities as she **can,to,decreased** isolation." |

We required a robust mechanism to identify and quantify the extent of errors (mentioned in Table 1) in the dataset. The ACH dataset contained various types of errors within a single data instance, making it difficult to use existing tools like Autocorrect [54], PyAspeller [55], and Grammar-check [56] for error quantification, as these tools typically handle one type of error at a time. Their performance, particularly in identifying spelling mistakes, depends on a predefined dictionary. However, in the case of medical records, many specialized medical terms may not exist in standard dictionaries used by such tools. Furthermore, many studies on improving text quality often emphasized grammatical errors [57–59]. Recently, LLMs have been applied to correct grammatical mistakes in textual datasets, making them the most advanced tools available for improving text quality[57, 59].

Many LLMs can be found in the literature [14], but considering the trade-offs between computational complexity, availability (open source) and performance [60, 61], we selected Mixtral 8x7B for this study [62]. It is important to note that a comprehensive evaluation of various LLMs for dataset correction is beyond the scope of this study.

#### 3.2.2. Quantification of the Errors

To quantify the extent of errors and assess their impact on prediction tasks, we tested Mixtral across different categories of errors (Table 1) to assess its performance. We conducted similar tests with GPT4. We found that the LLMs struggled to differentiate between various types of errors, such as distinguishing spelling mistakes from grammatical errors. For example, when asked to correct only grammatical mistakes, the LLM would also correct any spelling mistakes. To address the limitation of distinguishing between different types of errors, we developed a generalized metric for error quantification. This approach involved splitting each textual data instance into tokens based on white spaces and evaluating each token for errors. Hence, the error rate was defined as:

$$Error\ rate = \frac{tokens\ with\ errors}{total\ tokens} \tag{1}$$

This metric disregarded punctuation and grammatical errors, focusing instead on missing whitespace between words and spelling mistakes (Table 1), which was frequently observed in the ACH dataset. The error rate was computed on the instance level and the average was computed for the complete dataset.

This quantification was important as the study aimed to assess the impact of varying error rates on feature representation and machine learning models. The mentioned error rate was used to synthetically add errors to the MIMIC dataset (Section 3.2.4) and quantify the extent of errors found in the ACH dataset.

To quantify the error rate, the ACH data instances were tokenized on white-space, and then a query was designed for Mixtral to identify the tokens with errors. A sample query and Mixtral’s response are given in Supplementary Table S1. The response from Mixtral was utilized to quantify the error rate of the dataset, serving as a baseline for errors commonly found in low-quality real-world datasets.

It is well-known that the performance of LLMs is not perfect i.e., the outcome generated by these models may contain errors. In this case, Mixtral might detect an error that does not exist or miss an actual error in the data. To evaluate the accuracy of LLM error quantification, we randomly sampled 150 data instances from the ACH dataset. These 150 instances were manually corrected to create a high-quality, error-free dataset, referred to as the 'ground truth' for this study. The error rate was computed by Mixtral for this ground truth dataset using the aforementioned methodology. Analyzing these results helped in assessing Mixtral's performance in detecting errors from the given dataset.

#### 3.2.3. Data Correction

As previously mentioned in Section: 3.2.1, traditional tools and methods are not suitable for correcting text in cases involving out-of-dictionary words and missing white spaces between words. Therefore, we used Mixtral to correct these errors. The correction query employed in the study is provided in Supplementary Table S2. It should be noted that this query specifically addresses spelling mistakes to maintain consistency with the error quantification outlined in Section: 3.2.1. It should be noted that Mixtral corrected grammatical mistakes as well for many instances.

The ground-truth dataset was corrected using Mixtral. This correction was then qualitatively compared with the manually corrected data instances to assess the performance of Mixtral in correcting textual errors within the dataset, whose details are reported in Section 4.

### 3.2.4. Erroneous Data Generation

In order to assess the impact of textual dataset quality on feature representation and machine learning models, we required a controlled generation of textual data with varying levels of errors. To achieve this, we introduced artificial errors into the MIMIC dataset corresponding to varying error rates. The study focused on two types of errors: misspellings and missing white spaces between words. The proportion of these two error types was randomly determined based on the overall error percentage. For example, in a dataset with 10% errors, this percentage could be randomly split, such as 4% for misspelt words and 6% for missing spaces. This was calculated using Equation (1), where the number of erroneous tokens or words was computed based on the total number of tokens in a given textual data instance for a given error rate.

To generate misspelt versions of words, we used Mixtral to create incorrect counterparts. For introducing missing white-space errors, we placed more emphasis on removing spaces between words separated by punctuation marks, as this pattern was observed in real-life datasets (Table 2). To achieve this, the probability of sampling words containing punctuation marks was 50% higher than tokens without punctuation when randomly selecting tokens for removing white spaces.

## 3.3. Adverse Events and Data Sampling

We utilized different adverse events prediction for the two selected datasets to assess performance across a broad range of tasks, rather than focusing on a single specialized task. It is important to note that the sampled textual notes for each individual were concatenated to form a single input for the prediction model.

**MIMIC dataset**

The mortality prediction task was selected for the MIMIC dataset, utilizing progress notes recorded within the first 24 hours of ICU admission to predict patient outcomes. The task is binary, with the outcomes classified as 'alive' (represented by a value of ‘0’) and 'deceased' (represented by a value of ‘1’). Detailed information about the data sampling process, including inclusion criteria, preprocessing steps, and any filtering applied, can be found in [63].

**ACH dataset**

For the ACH dataset, we focused on predicting risks of depression and falls among elderly. According to the World Health Organization (WHO), depression is one of the most common mental disorders which affects approximately 7% of the world's population followed by dementia (5%) and anxiety (3.8%) [64]. WHO global report states that approximately 28-35% of people aged 65 and over fall each year increasing to 32-42% for those over 70 years of age [65]. As previously mentioned, the reason for choosing a different prediction task compared to the MIMIC dataset was to ensure heterogeneity in the tasks. The main objective is to evaluate performance on high- and low-quality textual datasets, rather than focusing on mortality prediction in hospital and aged care settings.

The residents suffering from depression were identified using their medical history, i.e., any resident who was clinically diagnosed as depressed was considered as a case '1' for binary classification and absence of depression as '0'. Fall incident reports found in ACH dataset were used to identify the residents at risk of a fall. It is common for a resident to experience multiple falls during their stay. Hence, we predicted fall risk for two cases: 1) prediction of first fall risk (no fall history), and 2) prediction of fall risk considering the fall history. The overall performance of feature representation and machine learning models in both cases was consistent. For brevity, the prediction of the first fall risk is reported here and the details of fall risk with fall history can be found in Supplementary Data.

A history of 30, 60, and 90 days (referred to as the tracking period) was selected for the prediction tasks when the fall or depression event was recorded for a resident (Supplementary Figure S1). For residents with no adverse event, data was randomly selected from their overall stay period. This ensured that no bias was introduced by restricting the timeline for the no-risk residents. Multiple data points were selected for individuals with multiple fall incidents. These data points corresponded to distinct progress notes, each linked to a fall incident occurring at different points in time. Similarly, multiple data points for no-fall cases were also sampled randomly.

## 3.4. Selection and Evaluation of Feature Representation and Machine Learning Models

### 3.4.1. Feature Representation

In the following sections, we will describe two methods to derive features, traditional term frequency-inverted document frequency (TF-IDF) and deep learning-derived embedding features.

## TF-IDF Features

Traditionally text has been split into tokens and the tokens have been used as features in machine learning algorithms. We used the following regular expression to split the text **(?u)\b[a-zA-Z]+\b**. After converting the tokens into lowercase, the term frequency (TF) per document and the inverted document frequency (IDF) were calculated for each token in a document (implemented using sci*kit-learn* python library*)*.

After computing TF-IDF features, after removing the stopwords, we further performed the following selection of tokens:

- Top tokens- TF-IDF (top): selected the top 5,000 most frequent tokens.
- Tokens with a minimum frequency threshold – TF-IDF (min): selected tokens that had a frequency greater than 5. It should be noted that this threshold could be changed, but we chose a frequency of 5 to remove the comparatively most unique words found in the corpus. Testing different threshold values is beyond the scope of the study.
- Raw tokens – TF-IDF (raw): No filtering is applied i.e., all raw tokens were used for the predictive tasks.

## Embedding Features

i. Skip-gram word2vec embeddings

For tokenization, we employed the same strategy used for generating TF-IDF features, which involved applying the specified regular expression followed by converting tokens to lowercase. The generated tokens were matched against a dictionary of the *word2vec-google-news-300* model [66], which was pre-trained using 3 million words and phrases using Google News and has a vector dimension of 300. If a token is not matched in the dictionary of the trained model, it is skipped and not considered as a feature. Feature vectors generated for each token in a textual data instance by word2vec model were concatenated and averaged to calculate an embedding feature for the data instance.

ii. BERT embeddings

BERT (Bidirectional Encoder Representations from Transformers) [67] has two advantages compared to word2vec. The first one is that the tokenization relies on word-piece tokenizer that splits words into sub-words, which might increase the matching of tokens to embeddings, unlike word2vec that ignores the out-of-vocabulary tokens. Secondly, word2vec generates static embedding i.e., each word has a single fixed vector. BERT on the other hand generates an embedding vector that considers the context for a word.

We evaluated various variants of BERT models such as Clinical BERT [68] and LongFormer [69], but there were several issues applying these models directly to our study. The length of tokens in the sampled progress notes in the ACH setting was larger than the tokens limit set by BERT (512 tokens) and Longformer (4,096 tokens). Truncating the progress notes to accommodate these limits resulted in poor performance of classification tasks. To address this limitation, an embedding vector (768 dimensions) for each token was computed using BERT model. The embedding vectors for tokens in a progress note were concatenated and averaged to calculate the final feature vector for the classification task. We used the BERT model pre-trained on clinical data [68] for the study.

#### 3.4.2. Machine Learning Models and Evaluation

Adverse event prediction was modelled as a classification problem, in which 0 suggests no risk of adverse event and 1 implies otherwise. We selected logistic regression, random forests, and XGBoost for the study. The default parameters of these models in the *scikit-learn* library were used for the experiments. There are numerous machine and deep learning models available, but an elaborative evaluation of these models is beyond the scope of the study. The goal of the analysis was to assess whether the quality of the dataset impacts the performance of the machine learning models. Hence, finding the model that is least or most affected by the data quality was not the study's primary goal. We specifically focused on the algorithms that could be linked with explainability as healthcare solutions often refrain to use the machine learning model as a black box. We utilized 2/3 of the data for training and 1/3 of the data for evaluation. The performance of the models was primarily evaluated using the area under the receiver operating curve (ROC-AUC),

## 4. Results

### 4.1. Descriptive Analysis

#### 4.1.1. MIMIC-III and ACH datasets

MIMIC-III contains anonymized data of 46,520 patients, having 58,976 admissions to the hospital. The details of the sampling strategy (inclusion and exclusion conditions) can be found in [63]. It should be noted that this study only focused on predicting mortality using the first 24-hour window after admission, where [63] assessed multiple prediction windows. The descriptive analysis of the sampled data used in the study is reported in Table 2.

The anonymized data from ACH consisted of 7181 residents with an average length of stay of 1.2 ± 2.2 years. The characteristics of the cohorts for both the fall and depression risk prediction tasks are reported in Table 2 and Supplementary Table S3. It is important to note that the sampling window and the length of stay of the residents influenced the number of sampled individuals. Additionally, a selection criterion was applied to the progress notes, where any instance with fewer than 21 recorded progress notes (assuming at least one progress notes per day, covering 70% of the days in a month) was considered incomplete and was excluded from the analysis. This resulted in an inconsistent number of samples across different adverse event prediction tasks presented in Tables 2 and S3.

**Table 2:** Characteristics of the selected residents for predicting mortality, depression and first fall. All values are represented as mean (standard deviation-SD) followed by proportion in the cohort (percentage - %) in blue colour. Positive cases represent death events for the MIMIC dataset and fall/depression for the ACH dataset

| **Characteristics** | **MIMIC** | **ACH** | | | | | |
|---|---|---|---|---|---|---|---|
| | **Mortality Risk -** | **Depression Risk – 30 days** | **Depression Risk – 60 days** | **Depression Risk – 90 days** | **Fall Risk – 30 days** | **Fall Risk – 60 days** | **Fall Risk – 90 days** |

| Total distinct residents/patients | | 35774 | 6336 | 6341 | 6342 | 4536 | 4415 | 4282 |
|---|---|---|---|---|---|---|---|---|
| Age (years) | | 62.5 (16.5) | 85.5 (8.49) | 85.51 (8.48) | 85.5 (8.53) | 84.28 (9.00) | 84.52 (8.79) | 84.91 (8.63) |
| Gender | Male | 26993 – 42.7% | 2287 - 36.1% | 2289 - 36.1% | 2290 - 36.1% | 2245 - 30.9% | 1954 – 32.8% | 1703 – 33.1% |
| | Female | 20148 - 57.3% | 4049 - 63.9% | 4052 - 63.9% | 4052 - 63.9% | 5019 - 69.1% | 3996 – 67.2% | 3442 – 66.9% |
| Positive cases – total | | 4221 – 11.8% | 3273 - 51.7% | 3275 - 51.6% | 3275 - 51.6% | 3917 - 53.9% | 3917 - 65.8% | 3917 - 76.1% |
| Positive cases per gender | Male | 2344 – 6.6% | 1645 - 26% | 1646 - 26% | 1646 - 26% | 1977 - 27.2% | 1977 - 33.2% | 1977 - 38.4% |
| | Female | 1877 – 5.2% | 1628 - 25.7% | 1629 - 25.7% | 1629 - 25.7% | 1940 - 26.7% | 1940 - 32.6% | 1940 - 37.7% |

### 4.1.2. Ground-truth and Processed Datasets

As mentioned in Section 3.2.2, we first assessed Mixtral's performance in quantifying the errors in the textual dataset. For this purpose, we randomly sampled 150 progress notes from the ACH dataset and removed progress notes having less than 5 tokens to ensure that very short progress notes are not included in the analysis. This exclusion was based on observations from the ACH dataset, which indicated that short progress notes did not contain any errors. The remaining 136 progress notes were manually assessed to quantify the extent of errors e.g., following the same procedure discussed in Section 3.2.2, the progress note was divided into tokens and then erroneously tokens were quantified. This served as the ground truth used to assess the performance of Mixtral for error quantification, whose results are reported in Figure 2.

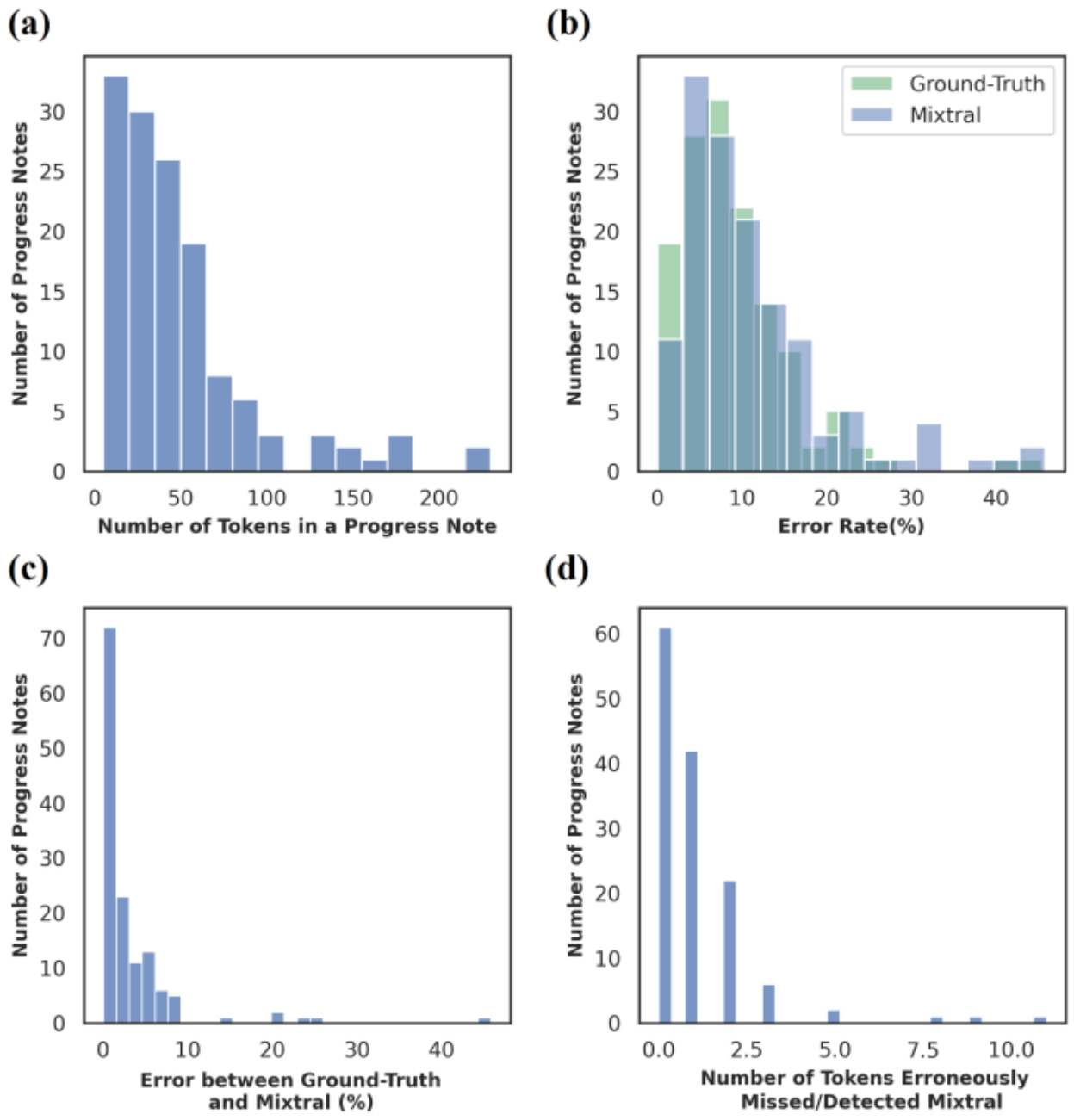

**Figure 2:** Mixtral's performance on 136 ground truth ACH progress notes for quantifying the textual errors (a) Number of tokens found in the ground truth progress notes. (b) Comparison of the error rate (%) found by Mixtral with the ground truth error rate. (c) The absolute error between the Mixtral computed and ground truth error rate. (d) The number of tokens that were erroneously detected or missed by Mixtral.

A performance deviation of 2.95%±5.67% (actual number of mismatched tokens 1.04 ±1.61) was observed for Mixtral from the ground-truth. The qualitative analysis of the results revealed that Mixtral erroneously identified the medical terminologies (e.g., ATOR, ADL, Perianal, situ, mane, obs, RN, etc.) as misspelt tokens. Moreover, a few cases of hallucinations were also observed e.g., Mixtral hallucinated misspelt "*angly*" from the following progress notes, "*Resident became angry during an activity*". Another example where Mixtral hallucinated "refuced" misspelled word from the following progress note, *"She refused to get out of bed and the staff tried many times to coax her out of bed to no avail."* This analysis demonstrated that Mixtral overestimated the errors for a few cases. Overall, Mixtral identified the correct misspelt tokens for 45% of the progress notes, and 17% of the progress notes had only one erroneously identified or missed token.

The error rate for the ACH dataset, considered poor-quality data (section 3.2.1), was found to be 7.5%±8.2% (Supplementary Figure S2). Considering this as a baseline of errors found in the real-world dataset, synthetic errors from 5% to 20% in increments of 5% were added to each progress note of the MIMIC dataset. This helped in the controlled simulation of poor-quality datasets, so we could systematically compare any change in the performance of the feature extraction and machine learning models with respect to the high-quality dataset (original MIMIC dataset in this case).

### 4.2. Feature Representation and Machine Learning Models

In this section, we present results comparing the performance of the machine learning classifiers trained on several feature representations of the progress notes with different levels of errors.

#### 4.2.1. Mortality Prediction

The results for the mortality prediction using different feature representation and machine learning models are presented in Figure 3. We found that the logistic regression had comparatively higher ROC-AUC than random forests and XGBoost for all feature representation models. In terms of the feature representation models, TD-IDF models outperformed neural network-based embedding models (word2vec and BERT).

Figure 3 clearly illustrates that, as hypothesized, the performance of the trained models declined as the level of errors in the dataset increased. A slight exception was observed with the XGBoost model for TF-IDF (raw).

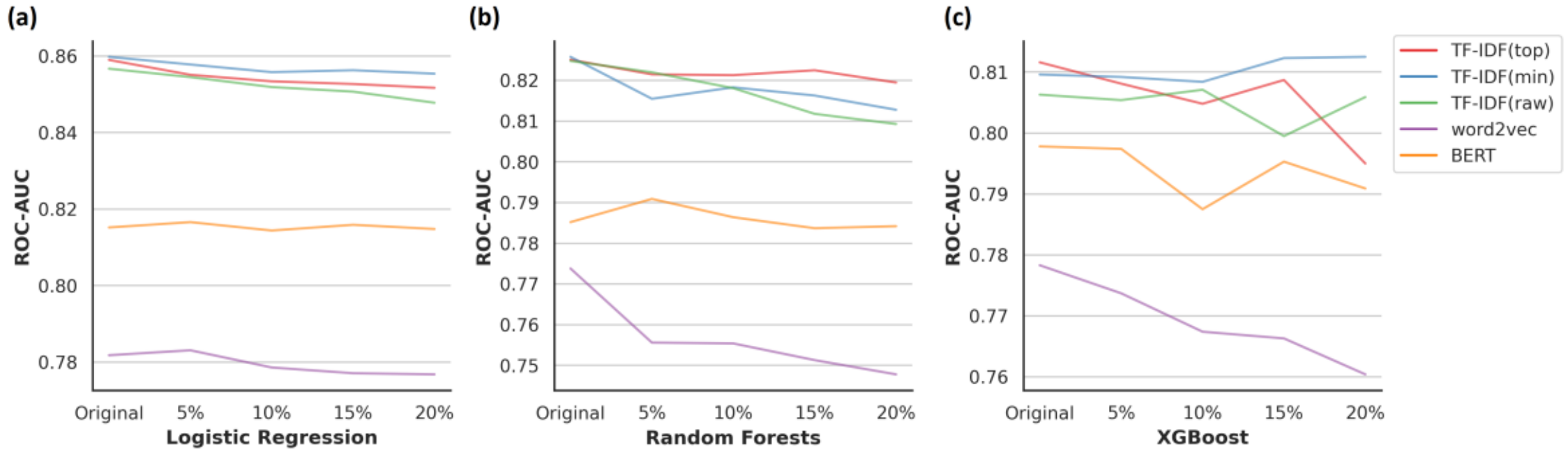

**Figure 3:** The performance of machine learning models for varying quality of MIMIC dataset using different feature representation techniques. TF-IDF (top): Top 5000 most frequent tokens; TF-IDF (min): Tokens with a minimum frequency of 5; TF-IDF (raw): All raw unfiltered tokens

#### 4.2.2. Depression and First Fall Prediction

The results of the prediction of depression and first fall for the real-world ACH dataset, are presented in Figures 4 and 5 respectively. A substantial difference between the original (low-quality progress notes with errors) and corrected (high-quality) dataset. This is consistent with the results of Section 4.2.1, where the substantial difference in decline of performance is observed for ≥10 error rate. It should be noted that the average error rate of ACH dataset is 7.5% (Section 4.1.2).

In terms of machine learning models, logistic regression had the lowest performance for the classification tasks. Similar to mortality prediction (Section 4.2.1), the traditional TF-IDF feature representation model outperformed the embedding models. The results for fall prediction with fall history were found consistent with these findings, whose details can be found in Supplementary Figure S3.

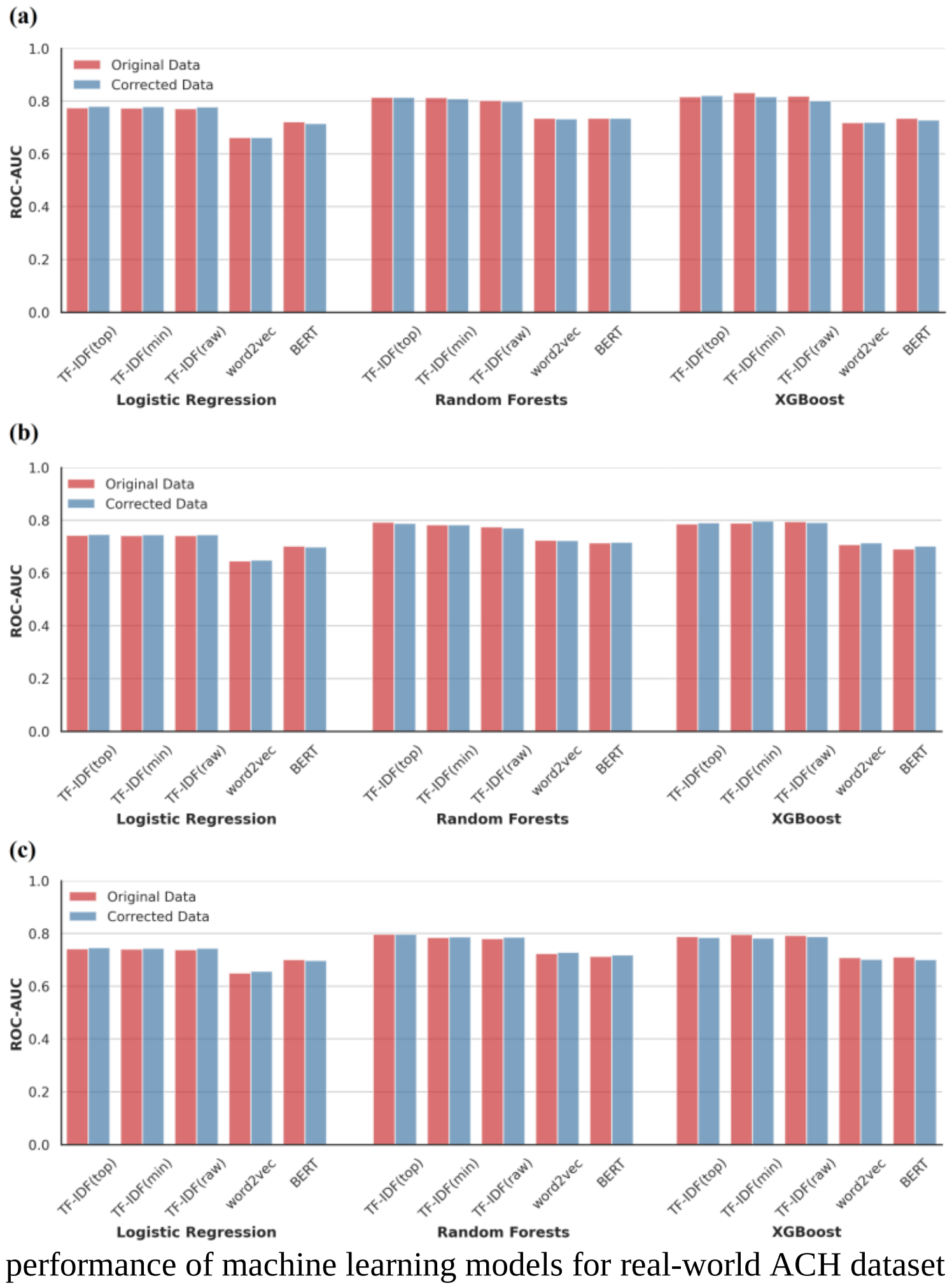

**Figure 4:** The performance of machine learning models for real-world ACH dataset using different feature representation techniques. The original and Mixtral's corrected progress notes were used for predicting **depression** under (a) 30 days, (b) 60 days, and (c) 90 days tracking period.

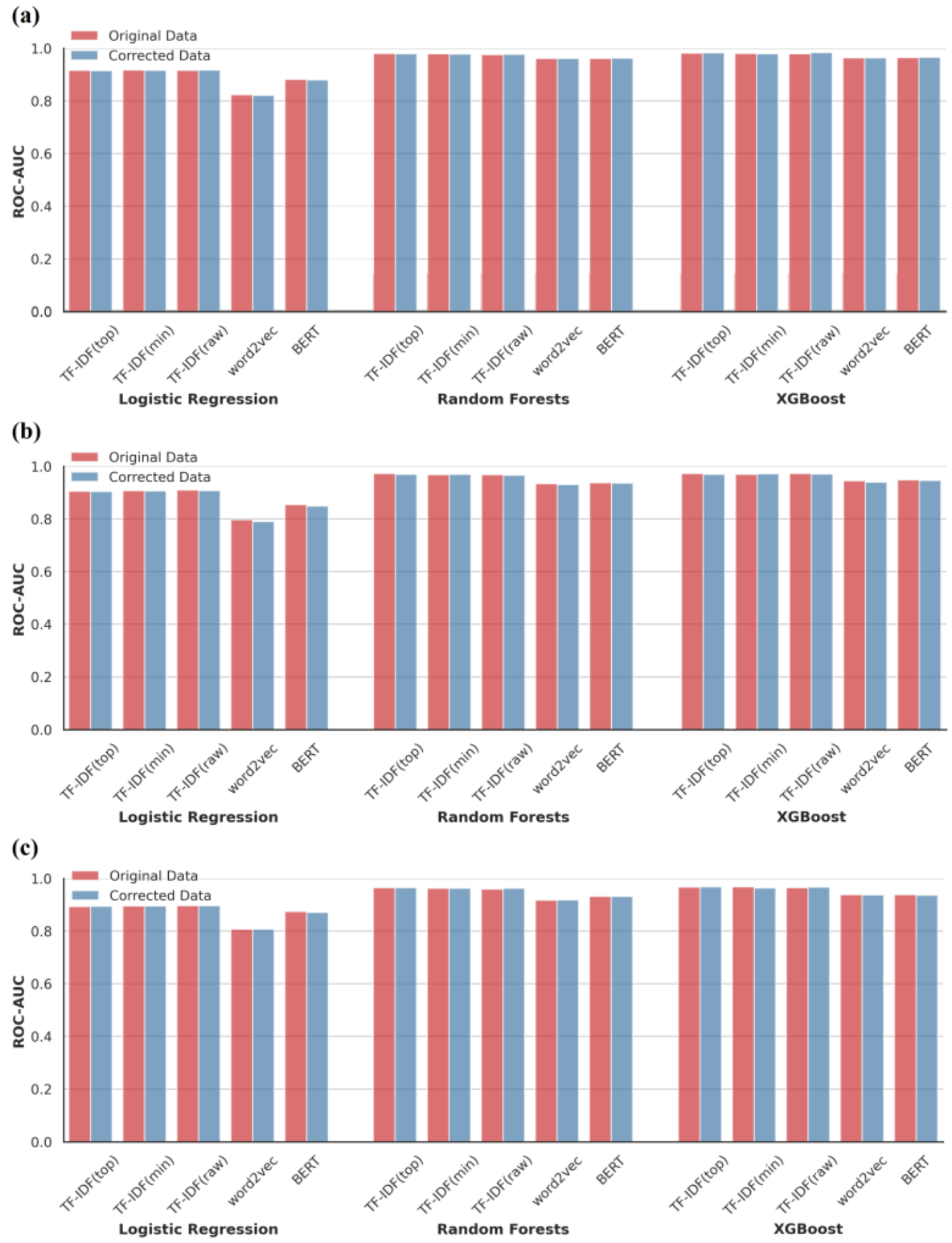

**Figure 5:** The performance of machine learning models for real-world ACH dataset using different feature representation techniques. The original and Mixtral's corrected progress notes were used for predicting **first fall** under (a) 30 days, (b) 60 days, and (c) 90 days tracking period.

## 5. Discussion

The performance of the machine learning models is dependent on the quality of the dataset used. Though the quality of the structured data has been extensively studied (Section 1), the study aimed to quantify the impact of the quality of unstructured textual data on the machine learning tasks. This is crucial, as it will demonstrate whether investing additional time and computational resources to enhance its quality is truly worthwhile.

Identifying errors in structured data is comparatively easier than in textual unstructured datasets due to its objective nature. However, the subjective and contextual nature of textual data makes defining and identifying different types of errors (Table 1) a challenging task. In Section 2, we discussed recent studies that have attempted to identify various quality issues in text, but this remains an open research question. Given its complexity, we focused on identifying errors at the token level (specifically, spelling mistakes and missing whitespace), instead of addressing errors at the sentence level (e.g., grammatical mistakes). This approach helped quantify the error rate as the ratio of the number of tokens with errors to the total number of tokens. This metric was crucial for evaluating the impact of data quality on the performance of feature representations and machine learning models

This study was designed to systematically assess the impact of textual dataset quality on feature representation and machine learning under two scenarios: using a (1) high-quality dataset and introducing varying levels of errors, and (2) low-quality dataset and correcting errors to improve its quality. For this purpose, we utilized the publicly available MIMIC-III high-quality dataset and an outsourced, real-world lower-quality dataset from ACHs. To introduce and correct errors, we employed Mixtral, as LLMs are increasingly being used to identify various issues in textual datasets (Section 2). While LLMs have demonstrated superior performance across many tasks [14, 60, 61], they are not perfect.

To evaluate Mixtral's performance in identifying errors, we created a manual ground-truth dataset of 136 progress notes from ACH dataset. This effort was necessary to ensure Mixtral's reliability in quantifying erroneous tokens, given that traditional dictionary-based tools often fail to detect errors such as missing whitespace between tokens which was frequently observed in the ACH dataset (Table 1). We found that Mixtral correctly detected errors in 63% of the progress notes, with 17% containing a single token mistakenly identified as an error. The qualitative analysis revealed that Mixtral misclassified medical terms as errors, leading to an overestimation of the error rate (Section 4.1.2). After evaluating Mixtral's performance, we used it to systematically introduce errors into the MIMIC-III dataset and to quantify the error rate of the ACH dataset (Section 4.1.2).

Also, we used Mixtral to correct errors in the ACH dataset to improve its quality. Unlike using the error rate at the token level, we provided the unprocessed progress notes as input for error correction (Supplementary Table S2). When assessing Mixtral's performance in error correction, we found that incorporating the context of the sentence led to better correction of errors compared to token-level corrections. Examples demonstrating Mixtral's poor performance on token-level error correction are provided in Supplementary Table S4. Sentence-level corrections not only resolved token-level errors but also addressed grammatical errors. While we acknowledge that grammatical mistakes were not a focus of this study, it is worth noting that no major grammatical issues have been reported in the MIMIC dataset. Therefore, the corrected ACH data also represented a high-quality dataset.

Tables S5 and S6 represent the impact of errors on the tokens extracted from the progress notes. Interestingly, as the error rate increased, the number of unique tokens also increased, while the average length of the notes decreased. This could be the result of the removal of white spaces and the addition of misspelt tokens to create a low-quality dataset.

We have used several methods to generate the features used for the machine learning methods. These features are derived from the text of the records in both datasets. Supplementary Table S7 shows the number of features in the MIMIC dataset depending on the feature representation and the error rate. It can be observed that the number of features increases as soon as the percentage of errors increases for the TF-IDF features. This was due to the unique erroneous tokens that increased with the error rate. For the raw features, i.e., with no filtering, there are 10 times more unique features compared to the original set, while it is 5 times in the case of the filtering of features by frequency.

Among the different machine learning models, logistic regression had comparatively the best performance for the mortality prediction task using the TF-IDF-derived features. It is evident from Figure 3, that the performance of the prediction models decreased with the increase in the error rate. Though the performance in terms of ROC-AUC is tolerant to small error rates (<10%), it starts to be significant with a high error rate (≥10%). From Figure 3, it could be observed that neural network-based embedding techniques with XGBoost are more sensitive to noise as their performance deteriorated for smaller error rates (<10%).

Supplementary Table S8 shows the statistics for the different partitions for the tasks for the ACH dataset. The number of unique features increases with the number of days, as a greater number of progress notes are generated over longer periods. It should be noted that despite the progress notes being longer, the proportion of additional unique words in the ACH progress notes was smaller compared to the MIMIC dataset. This might be due to the limited dictionary of ACH i.e., the smaller number of unique words irrespective of the length of the progress notes. This implies that the MIMIC dataset had a comparatively diverse dictionary. The ACH dataset includes daily notes from aged care residents, reports on activities and findings of the residents that might show more repetitions of similar progress notes as compared to the MIMIC notes. This might explain why we did not see a large difference with the corrected notes in terms of unique terms, which in turn might explain a limited impact on the performance of predictive tasks for the corrected ACH dataset (Figures 4-6). It should be noted that this is consistent with the finding of MIMIC dataset (Figure 3), where a major impact for many models was not observed for a <10% error-rate.

For the MIMIC dataset, linear regression shows higher prediction performance with TF-IDF features, which might indicate that individual words have higher predictive power for the mortality task. This seems to be consistent with the result of experiments with higher error rates as errors were introduced on the token level (Figure 3). We followed Mahbub et al. [63] work for predicting mortality, who also reported lower performance of deep learning-derived features. Among embedding-based techniques, we found that BERT outperformed word2vec (Figure 3). As mentioned in Section 3.4.1, word2vec ignores out-of-vocabulary words, which can lead to the loss of valuable information contained in these errors. BERT, on the other hand, addresses this limitation by subdividing words into sub-words and using these sub-words to generate embeddings. Consequently, it retains comparatively more information. Furthermore, the BERT model considers the context of each word within its surroundings. These properties likely contributed to BERT's superior performance compared to word2vec. We utilized pre-trained versions of word2vec and BERT, as retraining these models is computationally expensive. Additionally, retraining on erroneous datasets could lead the models to recognize erroneous tokens as valid, thereby nullifying the need to assess the true impact of errors on machine learning performance.

On the other hand, for the ACH dataset, the predictive performance of XGBoost was higher as compared to linear regression using TF-IDF features, which might indicate that feature correlation is important compared to individual features (Figures 4-6). It is well-established that no machine or deep learning model performs optimally across all tasks. This is also evident from our experiments, where the performance varied across different feature representations and machine learning models.

Overall, we found that the feature representation and machine learning models are tolerant of small errors (<10%). We found that the real-world ACH dataset consists errors (approx. 8%) but the performance of the prediction tasks can be optimized by careful selection of the feature representation and machine learning models. For the case, when the error rate is large (≥10%), corrective measures are mandatory to ensure the good performance of the predictive models.

### 5.1. Limitations

There are a few limitations to our study that should be acknowledged. First, the error quantification metric we used was rudimentary and did not account for the various types of textual errors listed in Table 1. Similarly, while Mixtral was used to correct errors, many of the corrected issues (such as grammatical and typological errors) were not included in the error rate quantification. As mentioned earlier, this represents a complex problem, and further research is needed to address these challenges, which is beyond the scope of this study.

Additionally, we tested only a limited selection of feature representations and machine learning models, without conducting a systematic, in-depth analysis of the wide range of models available in the literature or performing comprehensive parameter tuning. Also, our analysis primarily focused on the ROC-AUC evaluation metric, which is widely used for assessing the performance of machine learning tasks. Future work will include a detailed examination of various feature representations and machine learning models, along with a comprehensive analysis using different evaluation metrics.

In this study, we selected LLMs based on their computational feasibility and open-source availability. However, we did not evaluate their performance in correcting textual data. Extending this study to quantitatively evaluate different LLMs for data correction could provide valuable insights to the community in selecting LLMs that offer the best performance for improving the quality of textual data.

While embedding models such as BERT can directly process progress notes, they are constrained by a token limit of 512. Models with higher token limits, such as Longformer (4096 tokens), were also unsuitable, as many progress notes exceeded these limits. Consequently, we processed the progress notes at the token level. Further research is needed to develop methods that can effectively handle longer notes, which may include exploring methods to remove a subset of unimportant textual notes, a topic which is beyond the scope of this study.

## Conclusion

In this study, we evaluated the impact of low-quality textual data on feature representation and machine learning models. Unlike structured data, assessing the quality of textual data presents significant challenges due to the complex nature of many errors, such as grammatical, typographical, and missing whitespace mistakes. To address this, we proposed a rudimentary error evaluation metric focused on spelling and missing whitespace errors to assess the quality of two datasets: the real-world ACHs dataset and the open-source MIMIC dataset. We systematically introduced errors into the MIMIC dataset to analyze their impact on feature representations and machine learning models. Additionally, we leveraged large language models (LLMs), particularly Mixtral, for identifying, correcting, and corrupting the textual data. By comparing the performance of models on datasets with varying error rates (evaluated in terms of ROC-AUC), we observed that models were relatively tolerant to smaller error rates (<10%), but their performance significantly deteriorated at higher error rates (≥10%).In the real-world ACHs dataset, we measured an error rate of ≈8%, which did not result in a major decline in predictive performance. Therefore, for datasets with higher error rates, corrective measures are essential to ensure the reliability of machine learning models.

The study also revealed that traditional TF-IDF techniques outperformed embedding models for medical predictive tasks, underscoring the importance of selecting feature representations tailored to the problem domain. Similarly, the choice of machine learning models was dataset-dependent: logistic regression yielded better results for the MIMIC dataset, while tree-based ensemble methods performed better for the ACHs dataset.

## Abbreviations

AACQSC: Australian Aged Care Quality and Safety Commission
ACHs: Aged Care homes
BERT: Bidirectional Encoder Representations from Transformers
EHRs: Electronic Health Records
IDF: Inverted Document Frequency
LLMs: Large Language Models
NLP: Natural Language Processing
ROC- AUC: Area Under the Receiver Operating Curves
TF: Term Frequency
TF-IDF: Term Frequency-Inverted Document Frequency
WHO: World Health Organization

## Acknowledgement

This work was partially supported by Telstra Health and the Digital Health CRC Limited (DHCRC). DHCRC is funded under the Australian Commonwealth's Cooperative Research Centres (CRC) Program.
This research was supported by use of the Nectar Research Cloud, a collaborative Australian research platform supported by the NCRIS-funded Australian Research Data Commons (ARDC).

This research was supported by The University of Melbourne's Research Computing Services and the Petascale Campus Initiative.

## Conflict of Interest

We do not have any conflict of interest with respect to this study.

## Ethics approval and consent to participate

The Human Research Ethics Committee of RMIT University (Project 23257), Australia provided Ethics approval for the study. Informed consent was obtained from the administration and management of the aged care homes who have informed consent from the residents and/or their legal guardians.

## Author's Contribution

TS and AJJY were responsible for the conception, design, and data analysis of the study. LC reviewed the results and was involved in the interpretation of the data. TS and AJJY wrote the initial draft of the manuscript, and LC revised the manuscript critically for important intellectual content.

## Data Availability

MIMIC-III is an open-source dataset available at https://physionet.org/content/mimiciii/1.4/

The aged-care home dataset used and/or analysed during the current study cannot be made publicly available as the data was acquired from an aged care facility that did not provide consent for their data to be shared. Please contact Sarah Evans (aged.disability@health.telstra.com) to request the data from this study.

# Supplementary Data

**Table S1**: Sample queries demonstrating the extraction of tokens with errors using Mixtral. The blue text represents the base query used for all data instances, while the red text was replaced for different data instances

| Ind. | Example Queries | Mixtral Outcome |
|---|---|---|
| 1 | do any tokens in the following list have a spelling mistake, missing space or punctuation errors? generate a table with two columns, the first column represents only the tokens containing errors and the other column represents the corrected token: [‘The’, ‘resident’, ‘didnot', ‘eat’, ‘much’, ‘for’, ‘dinner’, ‘she’, ‘had’, ‘a’, ‘cup’, ‘tea’, ‘few’, ‘spoonful’, ‘of’, ‘soup’ , ‘and’, ‘main’, ‘meal.’] | Tokens with Errors \| Corrected Tokens<br>didnot \| did not<br>spoonful \| spoonfuls |
| 2 | do any tokens in the following list have a spelling mistake, missing space or punctuation errors? generate a table with two columns, the first column represents only the tokens containing errors and the other column represents the corrected token: [‘XYZ’, ‘is’, ‘also’, ‘walked’, ‘to’, ‘crafts,and’, ‘atttend’, ‘as’, ‘many’, ‘group’, ‘activities’, ‘as’, ‘she’, ‘can,too,decreased’, ‘isolation.’] | Tokens with errors \| Corrected tokens<br>crafts,and \| crafts and<br>atttend \| attend<br>as,she \| as she<br>can,to,decreased \| can, too, decrease<br>isolation. \| isolation |

**Table S2**: Sample queries demonstrating the data correction using Mixtral. The blue text represents the base query used for all data instances, while the red text was replaced for different data instances

| Ind. | Example Queries | Mixtral Outcome |
|---|---|---|
| 1. | Strictly proofread and correct any spelling mistakes in the following sentences. Only produce a single corrected text and do not provide any explanation. Return 'nil' if there is no mistake: “Resident buzzed at2300hrs on5/10/10,her legs felt like they were burning n she was in pain” | "The resident buzzed at 2300 hours on 5/10/10; her legs felt like they were burning, and she was in pain." (Corrected) |
| 2. | Strictly proofread and correct any spelling mistakes in the following sentences. Only produce a single corrected text and do not provide any explanation. Return 'nil' if there is no mistake: “Resident was sleeping on round check,repositioned by2 x staff fluids given. nil problems settled ator” | "The resident was sleeping during routine check; repositioned by two staff members. No fluid issues were settled or ator." |

**Table S3:** Characteristics of the selected residents for prediction of fall risk (instances contain residents with fall history). All values are represented as mean (standard deviation-SD) followed by proportion in the cohort (percentage -%) in blue colour.

| Characteristics | | Fall Risk – 30 days | Fall Risk – 60 days | Fall Risk – 90 days |
|---|---|---|---|---|
| Total distinct residents | | 4491 | 4412 | 4281 |
| Age (years) | | 85.15 (8.77) | 85.53 (8.56) | 85.64 (8.53) |
| Gender | Male | 3740 - 30.7% | 3457 - 28.4% | 3058 - 32% |
| | Female | 8426 - 69.3% | 7517 - 61.8% | 6501 - 68% |
| Deterioration Cases – total | | 6163 - 50.7% | 6163 - 50.7% | 6163 - 64.5% |
| Cases per gender | Male | 3168 - 26% | 3168 - 26% | 3168 - 33.2% |
| | Female | 2995 - 24.7% | 2995 - 24.7% | 2995 – 31.3% |

**Table S4**: Sample queries demonstrating the data correction using Mixtral on token level. The blue text represents the base query used for all data instances, while the red text was replaced for different data instances

| Ind. | Example Queries | Mixtral Outcome |
|---|---|---|
| 1. | Correct any spelling mistakes in the following list. Only generate a list containing all the provided tokens along with the ones that have been corrected do not provide any explanation: ["Resident", "buzzed", "at2300hrs", "on5/10/10,her", "legs", "felt", "like", "they", "were", "burning", "n", "she", "was", "in", "pain”] | ["Resident", "buzzed", "at 2300 hours", "on 5/10/10", "her legs", "felt like", "they were burning", "she was in", "pain"] |
| 2. | Correct any spelling mistakes in the following list. Only generate a list containing all the provided tokens along with the ones that have been corrected do not provide any explanation: ["Resident", "was", "sleeping", "on", "round", "check,repositioned", "by2", "x", "staff", "fluids", "given.", "nil", "problems", "settled", "ator"] | ["Resident", "was", "sleeping", "on", "the", "round", "check, repositioned", "by", "staff", "fluids", "given.", "nil", "problems", "were", "settled", "after", "the", "ator"] |

**Table S5:** Regular expression-based tokenizer statistics for different datasets and prediction tasks. This tokenization approach was used for TF-IDF and word2vec approaches

| Dataset | Prediction Task | Experiment Condition | Unique tokens | Avg length (std) | Max length | Min length |
|---|---|---|---|---|---|---|
| MIMIC III | Mortality | Original notes (no error) | 129,069 | 1,641.28 (2,372.09) | 29,715 | 4 |
| | | 5% error | 494,049 | 1,567.90 (2,278.57) | 28,300 | 4 |
| | | 10% error | 740,008 | 1,531.81 (2,229.66) | 27,693 | 4 |
| | | 15% error | 973,248 | 1,498.84 (2,188.69) | 27,478 | 3 |
| | | 20% error | 1,219,423 | 1,468.29 (2,152.54) | 27,227 | 3 |
| Aged Care Home (RFBI) | Depression | 30-days tracking - original | 59,535 | 2,578.13 (2,322.06) | 18,616 | 7 |
| | | 30-days tracking - corrected | 38,307 | 2,721.01 (2,446.99) | 19,373 | 8 |
| | | 60-days tracking - original | 74,904 | 4,037.24 (3,623.64) | 30,432 | 7 |
| | | 60-days tracking - corrected | 46,090 | 4,258.67 (3,817.11) | 32,178 | 8 |
| | | 90-days tracking - original | 84,023 | 5,070.59 (4,531.26) | 34,089 | 7 |
| | | 90-days tracking - corrected | 50,675 | 5,350.08 (4,778.42) | 36,141 | 8 |
| | First Fall | 30-days tracking - original | 48,743 | 1,922.27 (1,522.90) | 19,844 | 10 |
| | | 30-days tracking - corrected | 32,769 | 2,029.64 (1,603.87) | 20,287 | 11 |
| | | 60-days tracking - original | 60,443 | 3,110.62 (2,618.96) | 21,905 | 12 |
| | | 60-days tracking - corrected | 38,513 | 3,280.91 (2,757.69) | 22,827 | 12 |
| | | 90-days tracking - original | 66,593 | 4,117.92 (3,408.97) | 28,333 | 12 |
| | | 90-days tracking - corrected | 41,440 | 4,341.63 (3,589.90) | 29,486 | 12 |
| | Fall with Fall History | 30-days tracking - original | 63,814 | 1,760.24 (1,422.99) | 19,844 | 10 |
| | | 30-days tracking - corrected | 40,335 | 1,858.47 (1,499.21) | 20,287 | 11 |
| | | 60-days tracking - original | 81,808 | 3,001.53 (2,356.72) | 21,456 | 23 |
| | | 60-days tracking - corrected | 48,875 | 3,163.77 (2,477.03) | 22,672 | 23 |
| | | 90-days tracking - original | 91,450 | 4,140.68 (3,099.86) | 27,850 | 18 |
| | | 90-days tracking - corrected | 53,532 | 4,365.00 (3,262.56) | 29,385 | 20 |

**Table S6:** BERT tokenizer statistics for different datasets and prediction tasks

| Dataset | Prediction Task | Experiment Condition | Unique tokens | Avg length (std) | Max length | Min length |
|---|---|---|---|---|---|---|
| MIMIC III | Mortality | Original notes (no error) | 15,465 | 3,955.69 (5,634.82) | 66,833 | 9 |
| | | 5% error | 15,719 | 3,583.78 (5,074.80) | 61,179 | 9 |
| | | 10% error | 15,849 | 3,634.20 (5,153.00) | 61,934 | 9 |
| | | 15% error | 15,888 | 3,687.11 (5,241.47) | 63,952 | 8 |
| | | 20% error | 15,942 | 3,744.27 (5,341.49) | 66,215 | 10 |
| Aged Care Home (RFBI) | Depression | 30-days tracking - original | 14,236 | 3,618.31 (3,245.75) | 26,398 | 11 |
| | | 30-days tracking - corrected | 14,495 | 3,760.94 (3,356.32) | 27,327 | 11 |
| | | 60-days tracking - original | 14,924 | 5,658.85 (5,066.26) | 42,983 | 11 |
| | | 60-days tracking - corrected | 15,205 | 5,882.29 (5,243.26) | 44,600 | 11 |
| | | 90-days tracking - original | 15,227 | 7,110.86 (6,373.50) | 48,197 | 11 |
| | | 90-days tracking - corrected | 15,554 | 7,396.76 (6,608.38) | 50,184 | 11 |
| | First Fall | 30-days tracking - original | 13,675 | 2,628.43 (2,063.69) | 26,407 | 16 |
| | | 30-days tracking - corrected | 13,991 | 2,747.68 (2,139.08) | 26,235 | 17 |
| | | 60-days tracking - original | 14,293 | 4,229.56 (3,528.94) | 30,057 | 16 |
| | | 60-days tracking - corrected | 14,638 | 4,418.15 (3,666.33) | 31,065 | 17 |
| | | 90-days tracking - original | 14,530 | 5,591.90 (4,595.62) | 38,343 | 16 |
| | | 90-days tracking - corrected | 14,879 | 5,842.49 (4,779.39) | 39,828 | 17 |
| | Fall with Fall History | 30-days tracking - original | 14,423 | 2,399.27 (1,929.27) | 26,407 | 16 |
| | | 30-days tracking - corrected | 14,774 | 2,509.76 (2,001.56) | 26,235 | 17 |
| | | 60-days tracking - original | 15,061 | 4,073.35 (3,188.23) | 30,057 | 35 |
| | | 60-days tracking - corrected | 15,423 | 4,257.21 (3,305.07) | 31,065 | 34 |
| | | 90-days tracking - original | 15,347 | 5,618.83 (4,190.68) | 38,343 | 27 |
| | | 90-days tracking - corrected | 15,770 | 5,872.59 (4,353.04) | 39,828 | 29 |

**Table S7:** Number of features for different NLP feature representation models for mortality prediction task for the MIMIC dataset.

| Feature Representation | Number of features (dimensionality of input vector) | | | | |
|---|---|---|---|---|---|
| | Original | 5% error rate | 10% error rate | 15% error rate | 20% error rate |
| TF-IDF (top) | 5,000 | | | | |
| TF-IDF (min) | 32,804 | 84,962 | 119,875 | 148,736 | 175,915 |
| TF-TDF (raw) | 128,921 | 493,901 | 739,859 | 973,099 | 1,219,274 |
| word2vec | 300 | | | | |
| BERT | 786 | | | | |

**Table S8:** Number of features for different NLP feature representation models for the deterioration prediction tasks in the ACH setting

| Deterioration Prediction Tasks | Feature Representation | Number of features (dimensionality of input vector) | | | | | |
|---|---|---|---|---|---|---|---|
| | | 30-day tracking | | 60-day tracking | | 90-day tracking | |
| | | Original | Corrected | Original | Corrected | Original | Corrected |
| Depression | TF-IDF (min) | 14,626 | 13,208 | 18,007 | 15,631 | 20,042 | 17,003 |
| | TF-IDF (raw) | 59,388 | 38,156 | 74,756 | 45,939 | 83,873 | 50,523 |
| First Fall | TF-IDF (min) | 14,265 | 12,807 | 16,040 | 14,084 | 16,817 | 14,610 |
| | TF-IDF (raw) | 48,606 | 32,621 | 60,304 | 38,364 | 66,453 | 41,290 |
| Fall with history | TF-IDF (min) | 17,380 | 15,009 | 21,329 | 17,594 | 23,150 | 18,740 |
| | TF-IDF (raw) | 63,675 | 40,185 | 81,667 | 48,725 | 91,310 | 53,381 |
| Common across all the tasks | TF-IDF (top) | 5,000 | | | | | |
| | word2vec | 300 | | | | | |
| | BERT | 768 | | | | | |

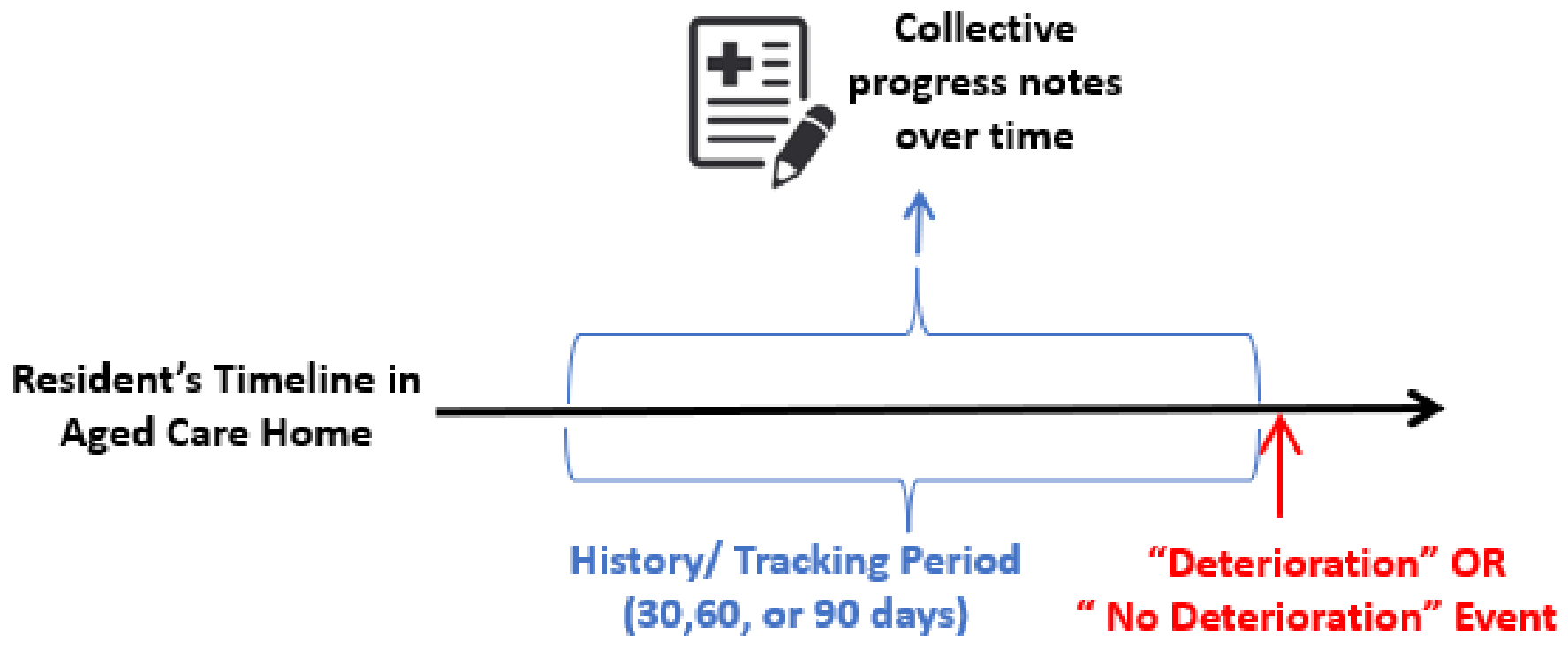

**Supplementary Figure S1:** Progress notes sampling process for the deterioration risk

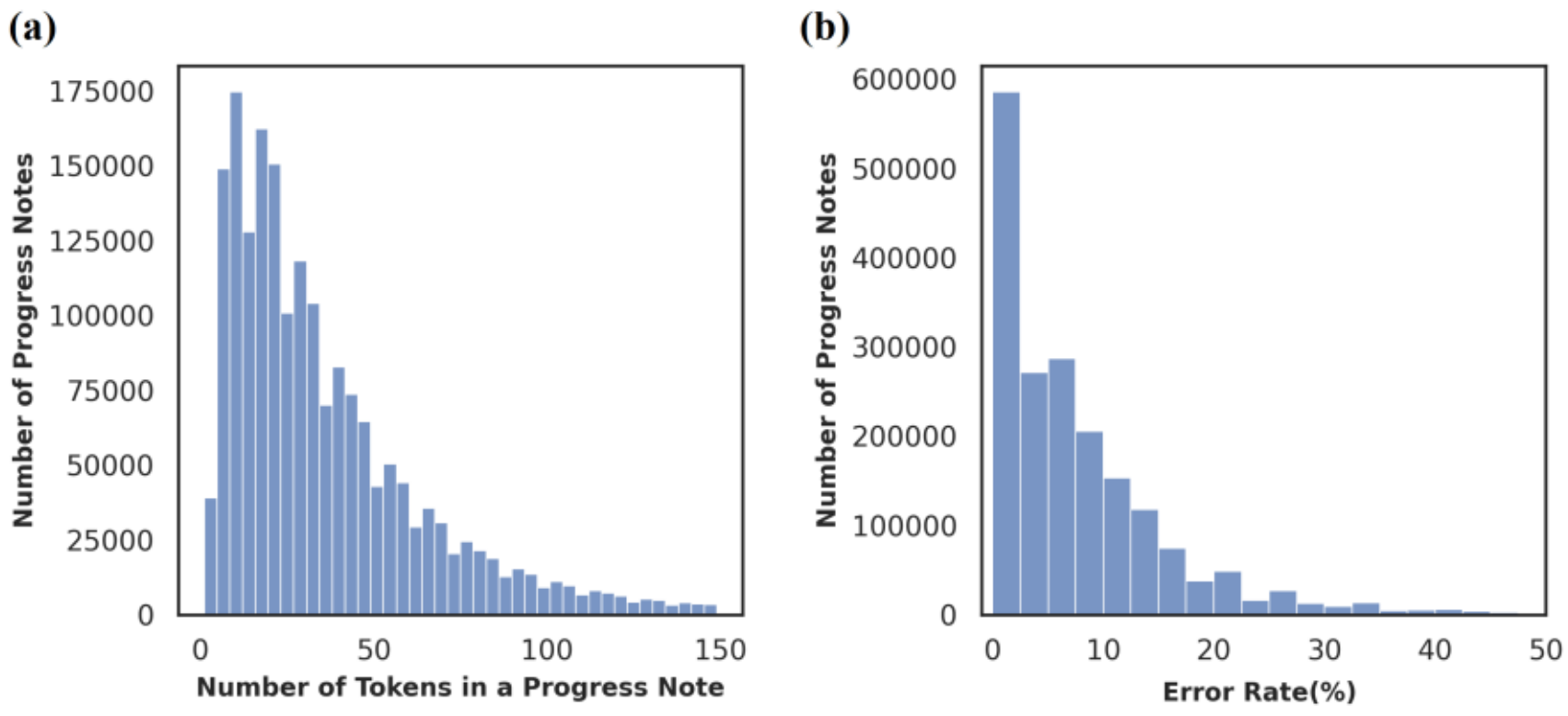

**Figure S2:** (a) The histogram demonstrating the number of tokens (length) in progress notes. RFBI dataset consisted of 1,903,936 progress notes. A threshold of <150 tokens was applied for clarity of the figure. It should be noted that 40,032 (2.1%) progress notes have a length greater than 150 tokens. (b) The extent of errors found in the RFBI dataset

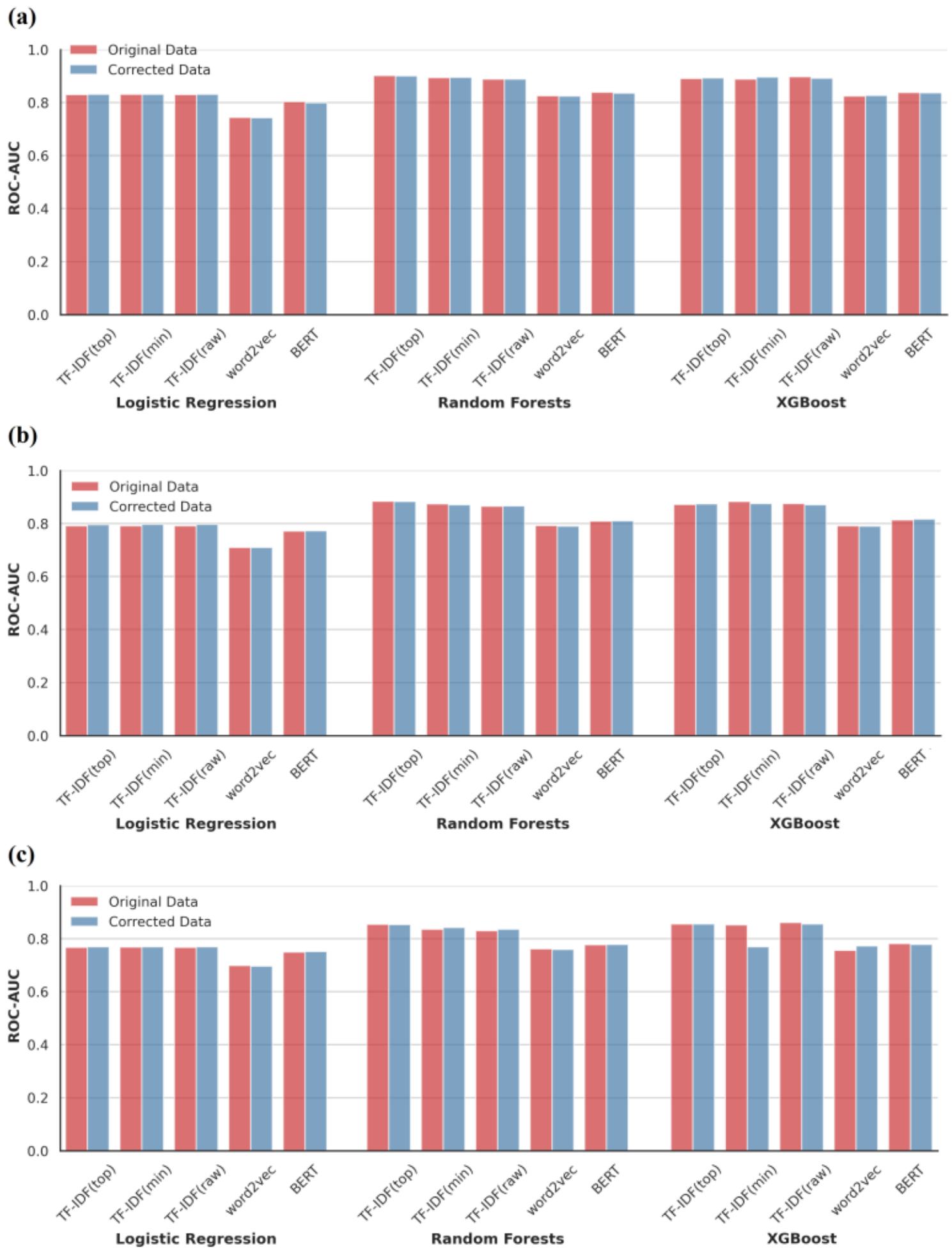

**Figure S3:** The performance of machine learning models for real-world RFBI dataset using different feature representation techniques. The original and Mixtral's corrected progress notes were used for predicting fall for residents with fall history under (a) 30 days, (b) 60 days, and (c) 90 days tracking period.